\documentclass{article}

\usepackage[final, nonatbib]{neurips_2019}

\usepackage[square,numbers]{natbib}
\usepackage[utf8]{inputenc} 
\usepackage[T1]{fontenc}    
\usepackage{hyperref}       
\usepackage{url}            
\usepackage{booktabs}       
\usepackage{amsfonts}       
\usepackage{nicefrac}       
\usepackage{microtype}      
\usepackage[pdftex]{graphicx}
\usepackage{textcomp}

\title{Modeling patterns of smartphone usage and their relationship to cognitive health}

\author{%
  Jonas Rauber\thanks{Work done during an internship at Apple.} \\
  University of T\"ubingen\\
  \texttt{jonas.rauber@bethgelab.org} \\
  \And
  Emily B. Fox \\
  Apple \\
  \texttt{emily\_fox@apple.com} \\
  \And
  Leon A. Gatys \\
  Apple \\
  \texttt{lgatys@apple.com} \\
}

\begin{document}

\maketitle

\begin{abstract}
\vspace{-0.05in}
The ubiquity of smartphone usage in many people's lives make it a rich source of information about a person's mental and cognitive state. 
In this work we analyze 12 weeks of phone usage data from 113 older adults, 31 with diagnosed cognitive impairment and 82 without.
We develop structured models of users' smartphone interactions to reveal differences in phone usage patterns between people with and without cognitive impairment.
In particular, we focus on inferring specific \emph{types} of phone usage sessions that are predictive of cognitive impairment.
Our model achieves an AUROC of 0.79 when discriminating between healthy and symptomatic subjects, and its interpretability enables novel insights into which aspects of phone usage strongly relate with cognitive health in our dataset.
\end{abstract}

\section{Introduction \& Related Work}
\vspace{-0.3em} 
The ubiquity of smartphone usage in many people's lives make it a rich source of information about a person's mental and cognitive state.
Here, we solely focus on app usage patterns and investigate to what extent they are informative about a person's cognitive health.
There has been significant past work on analyzing smartphone app usage patterns in general
(e.g., \cite{bohmer2011falling, girardello2010appaware, do2011smartphone, morrison2018large, farrahi2008daily}), including many studies that predict user behaviour and characteristics based on app usage (e.g., \cite{baeza2015predicting, chittaranjan2013mining, bati2018trust, wang2015smartgpa, wang2018tracking, bai2012will, Singh:2017:IIS:3171581.3134730, zhao2016discovering, murnane2016mobile}).
Most closely related to our work are prior studies that aim to predict cognitive health and abilities from smartphone usage \cite{dagum2018digital, chen2019developing, gordon2019app}.
Gordon et al. \cite{gordon2019app} analyzed the relationship between app usage and cognitive function in \emph{healthy} older adults.
Characteristics of app usage such as number of apps installed, the average app duration or app usage by hour of the day was shown to be informative of Cogstate Brief Battery \cite{maruff2009validity} test scores. 
Chen et al. \cite{chen2019developing} integrate a larger number of data sources, including passively sensed data (e.g., activity data, heart rate, phone usage, sleep data), survey responses (mood, energy), and apps testing specific psycho-motor functions (e.g., typing speed) into one model.
Using an ensemble of gradient boosted trees \cite{chen2016xgboost} on 1k hand-engineered features, they achieve an area under the receiver-operator curve (AUROC) of 0.77 discriminating between healthy and symptomatic subjects.
Inspired by these results, we address the same task as \cite{chen2019developing} while using only data sources related to app usage. 
Our model uses unsupervised learning to find different \emph{types} of interaction sessions in a user's app stream.
When combining the learned session types with supervised prediction of cognitive health, we obtain an AUROC of 0.79.

Through a number of ablation studies we demonstrate the importance of different model design decisions, including learned app embeddings, segmenting of the app stream into sessions, and clustering the sessions into session types.
Finally, the interpretable structure of our model reveals novel insights into what aspects of phone usage have a strong relationship with cognitive health in our dataset.
For example, we find that the relation between important apps such as \emph{Messages} and cognitive health completely changes depending on what other apps are used in the same session.  As such, the notion of sessions, and the learned structure of such session content, is critical to our performance; solely examining which apps are commonly used by an individual is not sufficient.

\section{Dataset}
We use a subset of the data collected in a 12-week feasibility study which monitored 31 people with clinically diagnosed cognitive impairment and 82 healthy controls in normal living conditions \cite{chen2019developing}.
The age range of the subjects was between 60 and 75 with a median age of 66 and 66\% of the subjects were female.
In particular, we analyze the app usage event streams that consist of the timestamps and app identity of all app openings and closings for each user over the course of the study. 
Furthermore, we use the phone unlock/lock event streams that consist of the time-stamps for all phone unlock and lock events.
Overall this data amounts to more than 800k app launches and 230k phone unlock events.
More elaborate details on the study design, data collection and the full dataset can be found in \cite{chen2019developing}.


\section{Model Description}\label{Sec:Model}
\begin{figure}[t!]\label{Fig:Model}
    \includegraphics[width=1\linewidth]{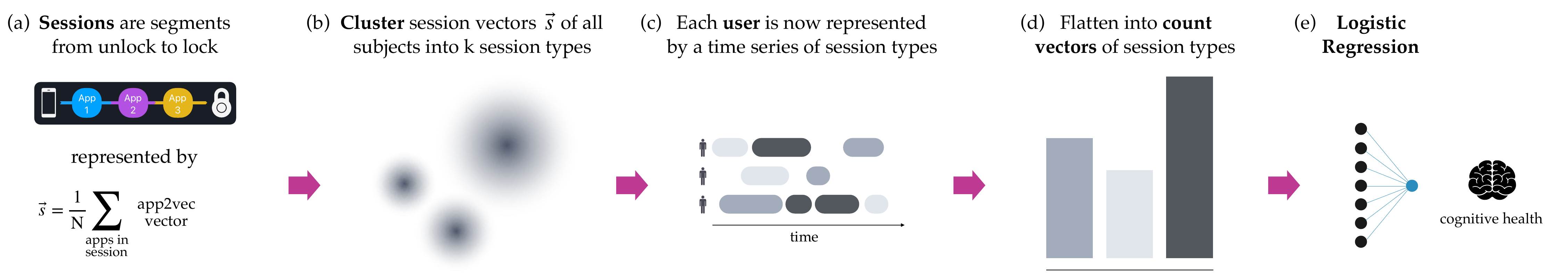}
    \caption{Overview of model architecture.}
    \vspace{-0.2in}
    \label{fig:model_desc}
\end{figure}



Our model first segments the app event stream of each user into a stream of \emph{interaction sessions} using the phone unlock/lock event stream (Fig.~\ref{fig:model_desc}a).
Thus all apps that are opened in between a pair of phone unlock and lock events are grouped into the same session.
To represent the many different apps in the dataset in a way that encodes similarity between apps, we train a 50-dimensional embedding in the same way as the popular word2vec \cite{mikolov2013efficient} by considering each user to be a ``sentence'' and predicting each app from the three apps before and after it in time.
To obtain a single vector representation \( \vec{s} \) of each session, we average the embeddings of all apps within each session.

Next, we use k-means to cluster all session vectors in the dataset to identify different \emph{session types} (Fig.~\ref{fig:model_desc}b).
A user's app usage is then represented by a time-series of session types (Fig.~\ref{fig:model_desc}c).


Finally, we summarize the time series of each user by counting the session types (Fig.~\ref{fig:model_desc}d).
For each user, we normalize the absolute session counts by the number of days the user participated in the study. In addition, we rescale all features of all users together such that the overall mean is 1 (to make the size of the features independent of the number of clusters). These features are then used as input to an L1-regularized logistic regressor to classify users as \emph{healthy} or \emph{symptomatic} (Fig.~\ref{fig:model_desc}e).

\section{Experiments \& Results}
Since the number of users, N, in our dataset is small, we perform our experiments using N leave-one-out (LOO) train/test splits.
For each of the N splits we select model hyper-parameters via a second LOO cross-validation loop on the N-1 training subjects.
The model parameters consist of the logistic regression weights and the hyper-parameters consist of (i) the number of session types, K, used for the session clustering and (ii) the inverse regularization strength, C, for the logistic regression. 
We evaluate the final performance by computing AUROC using the predicted probabilities from each of the N left out test subjects \cite{Bradley:1997:UAU:1746432.1746434} (Table \ref{tab:Ablation}).
Our full model achieves a test AUROC of 0.79, which is slightly higher than the 0.77 reported in \cite{chen2019developing} that used a much larger range set of input features.\footnote{Note though that \cite{chen2019developing} uses random 70/30 train/test splits instead of LOO for evaluation. This may limit their performance given the small size of the dataset.}

As described in Section~\ref{Sec:Model}, our full model entails segmenting the app stream into sessions, embedding apps into a vector space and averaging them to get session vectors, and clustering session vectors into session types.
Here, we systematically evaluate the impact of each of these model design choices on our ability to predict cognitive health using ablation studies.

Our first baseline (B1) tests the importance of grouping the app event stream into interaction sessions using the unlock/lock stream.
Instead of aggregating and clustering phone usage at the session level, we cluster the individual app embeddings directly.
We observe that performance drops from 0.79 to 0.75 when not grouping the app event stream in terms of sessions.

Our next three baselines (B2, B3, B4) aim to isolate the effect of the learned app embeddings.
In B2, we randomly permute the assignment between apps and their embeddings and find that performance drops from 0.79 to 0.69.
In B3 and B4, we replace the learned app embeddings with one-hot vectors encoding the app identity (B3) or coarser-scale App Store category (B4) of the app.
Session vectors are obtained by averaging the one-hot app vectors and a user is represented as the sum over session vectors instead of counts over learned session types.
While both B3 (0.75) and B4 (0.61) perform worse than our full model, the much larger drop for B4 indicates that App Store categories do not retain sufficient information to support the down-stream classification.


Our final two baselines (B5, B6) have the least structure, using one-hot encodings instead of learned embeddings (like B3 and B4) and no session aggregation (like B1).
Each user is represented as a vector of counts of the different apps (B5) or App Store categories (B6).
Again, we find that performance decreases (B3$\to$B5: 0.75$\to$0.72, B4$\to$B6: 0.61$\to$0.53) when not grouping the app event stream into sessions.

\begin{table}
\caption{Comparison to baseline models.}
\label{tab:Ablation}
\centering
\begin{tabular}{ ll cccccccc }
\toprule
Model && Full Model & B1 & B2 & B3 & B4 & B5 & B6 & Chance\\
\midrule
Test AUROC && \bf{0.79} & 0.75 & 0.69 & 0.75 & 0.61 & 0.72 & 0.53 & 0.50\\
\bottomrule
\end{tabular}
\vspace{-0.5em}
\end{table}%

\vspace{-0.05in}
\section{Model Introspection}
\vspace{-0.08in}

\begin{figure}
    \includegraphics[width=1\linewidth]{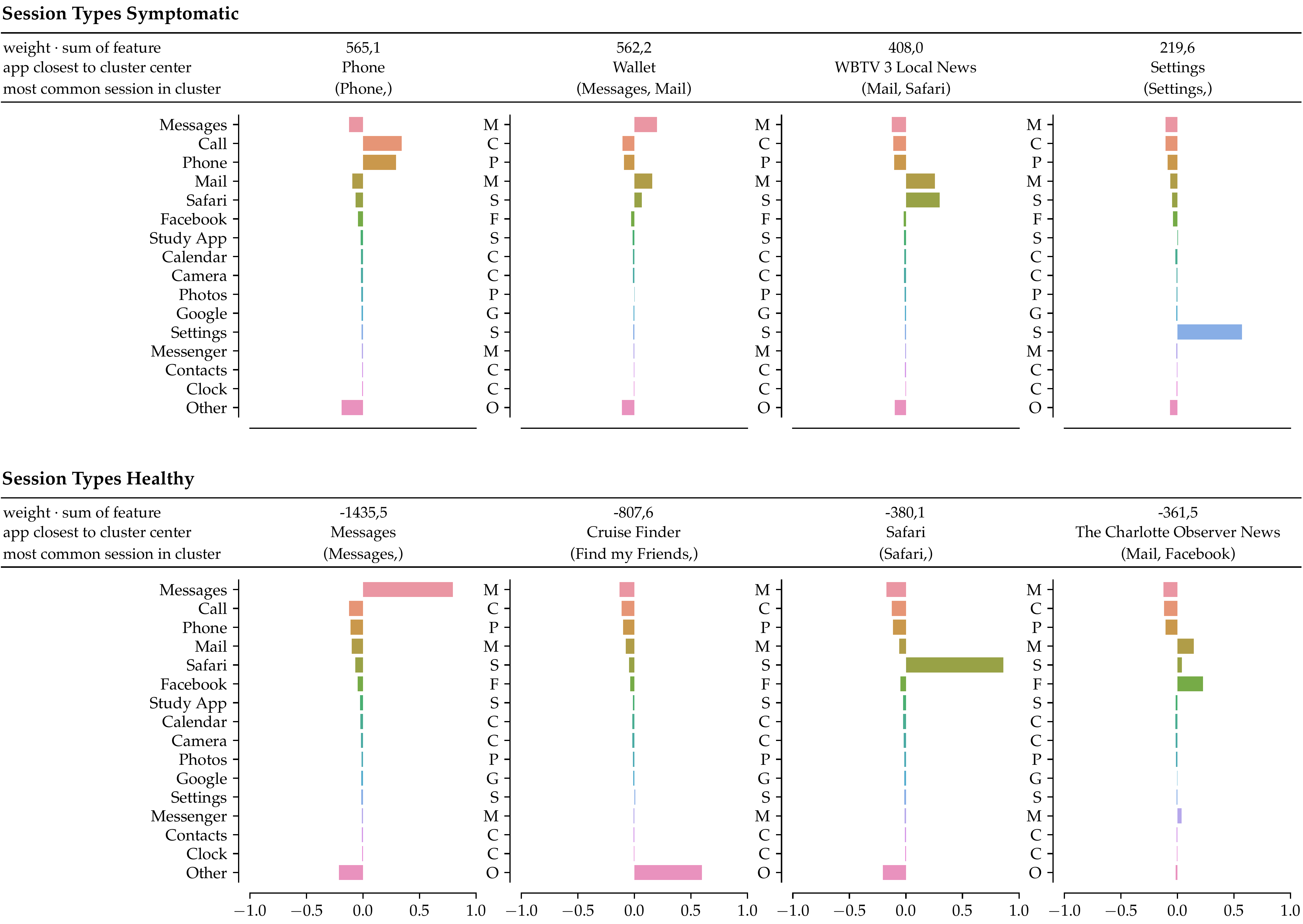}
    \caption{Analysis of session types with highest impact on classification.}
    \label{fig:model_intro}
    \vspace{-0.1in}
\end{figure}

\begin{figure}[h!]
    \includegraphics[width=1\linewidth]{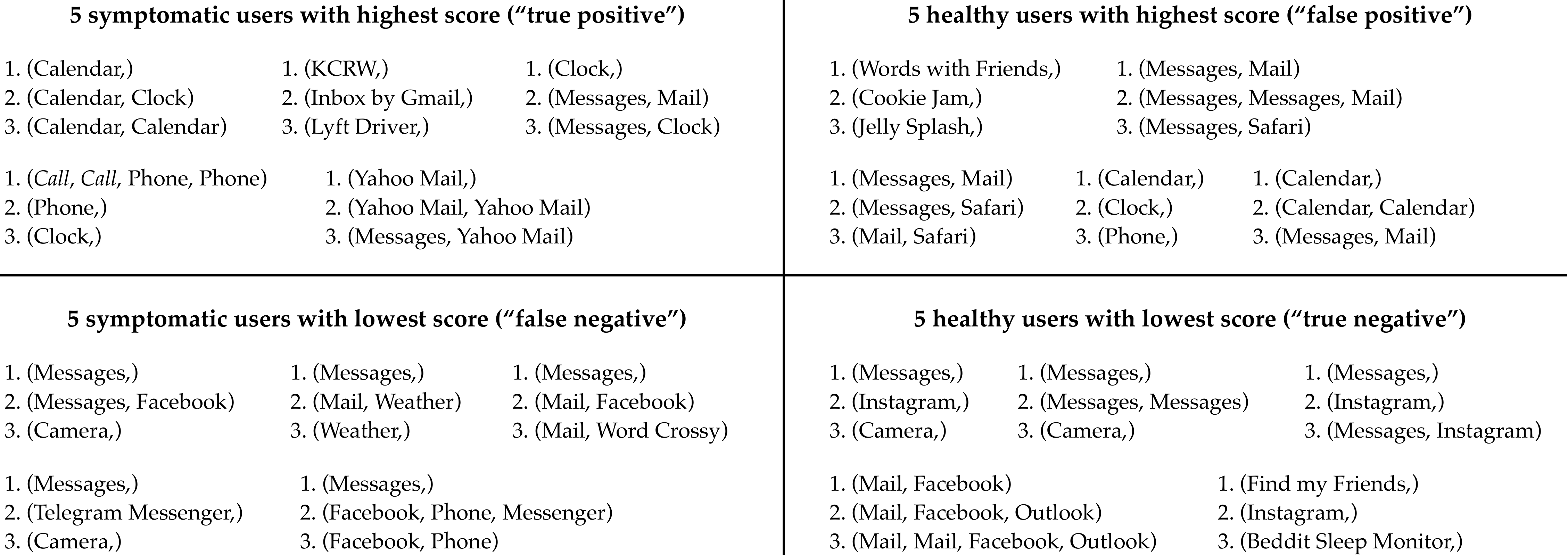}
    \caption{Twenty different users, five for each combination of \textit{healthy} (right) or \textit{symptomatic} (left) and \textit{high score} (top) or \textit{low score} (bottom). For each of these users, we list the three sessions with the largest contribution to the respective high or low score.}
    \label{fig:exp_dec}
    \vspace{-.5em}
\end{figure}

For our first analysis, we fit the model to all N subjects and analyse the four session types with the highest contribution to the model decision in either direction (Fig.~\ref{fig:model_intro}).
The contribution of a session type is measured by the product between the regression weight and the corresponding feature's value summed over all subjects (\emph{weight $\cdot$ sum of feature}).
To characterize each session type we report the app closest to the cluster center
\footnote{As described in section \ref{Sec:Model}, each session type is a cluster in the app2vec embedding space.} 
and the most common session in the session type.
Furthermore, for the 15 most common apps in the dataset, we visualize the difference between the app distribution in each session type and the overall distribution of apps in the dataset (Fig.~\ref{fig:model_intro}, bar plots).
The four session types that are most strongly associated with a high symptomatic score are dominated by \emph{Call} and \emph{Phone}, \emph{Messages} and \emph{Mail}, \emph{Mail} and \emph{Safari}, and \emph{Settings} (Fig.~\ref{fig:model_intro} upper half), followed by \emph{Clock} and \emph{Calendar} (not shown).
Session types most strongly related to a low score for symptomatic are dominated by \emph{Messages}, \emph{Safari}, \emph{Mail} and \emph{Facebook} and one session type consisting of many less frequent apps and thus difficult to summarize.
We observe a very interesting dependency of the influence of a session type on the interplay between multiple apps.
The session types dominated by \emph{Messages} and \emph{Mail} or \emph{Mail} and \emph{Safari} strongly increase the model's predicted score for symptomatic, whereas session types dominated by single \emph{Messages} or single \emph{Safari} sessions or by \emph{Mail} and \emph{Facebook} strongly decrease it.

To better understand our model's prediction for individual subjects, we use the N models resulting from the LOO procedure and analyze which sessions  cause them to (mis-)classify the respective test subjects.
Inference in our model is linear and thus we know the contribution of each session of a user to the model's prediction. In Fig.~\ref{fig:exp_dec} we show twenty different users, five for each combination of \textit{healthy} (right) or \textit{symptomatic} (left) and \textit{high score} (top) or \textit{low score} (bottom). For each of these users, we list the three sessions with the largest contribution to the respective high or low score.
For subjects with a high score (top) the most contributing single app sessions contain \emph{Phone}, \emph{Calendar} and \emph{Clock}.
For subjects with a low score (bottom) the most contributing single app sessions contain \emph{Messages}, \emph{Instagram} and \emph{Camera}.
As in our first analysis, we see that the impact of apps such as \emph{Messages} or \emph{Mail} strongly depends on the surrounding apps in the session. 
When \emph{Messages} shares a session with \emph{Mail} or \emph{Safari} it strongly increases the predicted score.
When \emph{Messages} is alone or in a session with \emph{Facebook} or \emph{Instagram} it strongly decreases the predicted score.
Overall we find that very similar sessions cause the model to correctly assign a high score to symptomatic subjects as well as to incorrectly assign a high score to healthy subjects (Fig.~\ref{fig:exp_dec} upper half) and similarly for (in-)correctly assigning low scores (Fig.~\ref{fig:exp_dec} lower half).

\section{Discussion \& Future Work}
\vspace{-0.2em}
The reported results have several potential limitations.
For example, the generalization of our results to the general population will be limited by size of the dataset and the fact that symptomatic subjects were already diagnosed when entering the study.

Nevertheless, it is exciting that app usage alone captures systematic differences between healthy and symptomatic subjects and we are actively pursuing multiple avenues to extend our model.
There are multiple parts in our model that can be replaced by more complex building blocks.
For example, one could use topic models \cite{blei2003latent} to extract session types or replace the logistic regression by a non-linear classifier such as gradient boosted trees or neural networks.
Additionally, we are aiming to incorporate the ordering of the apps in each session as well as user context such as time of the day or a user's motion state into the session representation.
Finally, we are exploring methods to learn the extraction of session types jointly with the classification of cognitive health in and end-to-end fashion.

\small

\bibliographystyle{abbrvnat}
\bibliography{references}

\begin{thebibliography}{22}
\providecommand{\natexlab}[1]{#1}
\providecommand{\url}[1]{\texttt{#1}}
\expandafter\ifx\csname urlstyle\endcsname\relax
  \providecommand{\doi}[1]{doi: #1}\else
  \providecommand{\doi}{doi: \begingroup \urlstyle{rm}\Url}\fi

\bibitem[Baeza-Yates et~al.(2015)Baeza-Yates, Jiang, Silvestri, and
  Harrison]{baeza2015predicting}
R.~Baeza-Yates, D.~Jiang, F.~Silvestri, and B.~Harrison.
\newblock Predicting the next app that you are going to use.
\newblock In \emph{Proceedings of the Eighth ACM International Conference on
  Web Search and Data Mining}, pages 285--294. ACM, 2015.

\bibitem[Bai et~al.(2012)Bai, Xu, Ma, Sun, and Zhao]{bai2012will}
Y.~Bai, B.~Xu, Y.~Ma, G.~Sun, and Y.~Zhao.
\newblock Will you have a good sleep tonight?: sleep quality prediction with
  mobile phone.
\newblock In \emph{Proceedings of the 7th International Conference on Body Area
  Networks}, pages 124--130. ICST (Institute for Computer Sciences,
  Social-Informatics and Telecommunications Engineering), 2012.

\bibitem[Bati and Singh(2018)]{bati2018trust}
G.~F. Bati and V.~K. Singh.
\newblock “trust us”: Mobile phone use patterns can predict individual
  trust propensity.
\newblock In \emph{Proceedings of the 2018 CHI Conference on Human Factors in
  Computing Systems}, page 330. ACM, 2018.

\bibitem[Blei et~al.(2003)Blei, Ng, and Jordan]{blei2003latent}
D.~M. Blei, A.~Y. Ng, and M.~I. Jordan.
\newblock Latent dirichlet allocation.
\newblock \emph{Journal of machine Learning research}, 3\penalty0
  (Jan):\penalty0 993--1022, 2003.

\bibitem[B{\"o}hmer et~al.(2011)B{\"o}hmer, Hecht, Sch{\"o}ning, Kr{\"u}ger,
  and Bauer]{bohmer2011falling}
M.~B{\"o}hmer, B.~Hecht, J.~Sch{\"o}ning, A.~Kr{\"u}ger, and G.~Bauer.
\newblock Falling asleep with angry birds, facebook and kindle: a large scale
  study on mobile application usage.
\newblock In \emph{Proceedings of the 13th international conference on Human
  computer interaction with mobile devices and services}, pages 47--56. ACM,
  2011.

\bibitem[Bradley(1997)]{Bradley:1997:UAU:1746432.1746434}
A.~P. Bradley.
\newblock The use of the area under the roc curve in the evaluation of machine
  learning algorithms.
\newblock \emph{Pattern Recogn.}, 30\penalty0 (7):\penalty0 1145--1159, July
  1997.
\newblock ISSN 0031-3203.
\newblock \doi{10.1016/S0031-3203(96)00142-2}.
\newblock URL \url{http://dx.doi.org/10.1016/S0031-3203(96)00142-2}.

\bibitem[Chen et~al.(2019)Chen, Jankovic, Marinsek, Foschini, Kourtis,
  Signorini, Pugh, Shen, Yaari, Maljkovic, et~al.]{chen2019developing}
R.~Chen, F.~Jankovic, N.~Marinsek, L.~Foschini, L.~Kourtis, A.~Signorini,
  M.~Pugh, J.~Shen, R.~Yaari, V.~Maljkovic, et~al.
\newblock Developing measures of cognitive impairment in the real world from
  consumer-grade multimodal sensor streams.
\newblock In \emph{Proceedings of the 25th ACM SIGKDD International Conference
  on Knowledge Discovery \& Data Mining}, pages 2145--2155. ACM, 2019.

\bibitem[Chen and Guestrin(2016)]{chen2016xgboost}
T.~Chen and C.~Guestrin.
\newblock Xgboost: A scalable tree boosting system.
\newblock In \emph{Proceedings of the 22nd acm sigkdd international conference
  on knowledge discovery and data mining}, pages 785--794. ACM, 2016.

\bibitem[Chittaranjan et~al.(2013)Chittaranjan, Blom, and
  Gatica-Perez]{chittaranjan2013mining}
G.~Chittaranjan, J.~Blom, and D.~Gatica-Perez.
\newblock Mining large-scale smartphone data for personality studies.
\newblock \emph{Personal and Ubiquitous Computing}, 17\penalty0 (3):\penalty0
  433--450, 2013.

\bibitem[Dagum(2018)]{dagum2018digital}
P.~Dagum.
\newblock Digital biomarkers of cognitive function.
\newblock \emph{npj Digital Medicine}, 1\penalty0 (1):\penalty0 10, 2018.

\bibitem[Do et~al.(2011)Do, Blom, and Gatica-Perez]{do2011smartphone}
T.~M.~T. Do, J.~Blom, and D.~Gatica-Perez.
\newblock Smartphone usage in the wild: a large-scale analysis of applications
  and context.
\newblock In \emph{Proceedings of the 13th international conference on
  multimodal interfaces}, pages 353--360. ACM, 2011.

\bibitem[Farrahi and Gatica-Perez(2008)]{farrahi2008daily}
K.~Farrahi and D.~Gatica-Perez.
\newblock Daily routine classification from mobile phone data.
\newblock In \emph{International Workshop on Machine Learning for Multimodal
  Interaction}, pages 173--184. Springer, 2008.

\bibitem[Girardello and Michahelles(2010)]{girardello2010appaware}
A.~Girardello and F.~Michahelles.
\newblock Appaware: Which mobile applications are hot?
\newblock In \emph{Proceedings of the 12th international conference on Human
  computer interaction with mobile devices and services}, pages 431--434. ACM,
  2010.

\bibitem[Gordon et~al.(2019)Gordon, Gatys, Guestrin, Bigham, Trister, and
  Patel]{gordon2019app}
M.~L. Gordon, L.~Gatys, C.~Guestrin, J.~P. Bigham, A.~Trister, and K.~Patel.
\newblock App usage predicts cognitive ability in older adults.
\newblock In \emph{Proceedings of the 2019 CHI Conference on Human Factors in
  Computing Systems}, page 168. ACM, 2019.

\bibitem[Maruff et~al.(2009)Maruff, Thomas, Cysique, Brew, Collie, Snyder, and
  Pietrzak]{maruff2009validity}
P.~Maruff, E.~Thomas, L.~Cysique, B.~Brew, A.~Collie, P.~Snyder, and R.~H.
  Pietrzak.
\newblock Validity of the cogstate brief battery: relationship to standardized
  tests and sensitivity to cognitive impairment in mild traumatic brain injury,
  schizophrenia, and aids dementia complex.
\newblock \emph{Archives of Clinical Neuropsychology}, 24\penalty0
  (2):\penalty0 165--178, 2009.

\bibitem[Mikolov et~al.(2013)Mikolov, Chen, Corrado, and
  Dean]{mikolov2013efficient}
T.~Mikolov, K.~Chen, G.~Corrado, and J.~Dean.
\newblock Efficient estimation of word representations in vector space.
\newblock \emph{arXiv preprint arXiv:1301.3781}, 2013.

\bibitem[Morrison et~al.(2018)Morrison, Xiong, Higgs, Bell, and
  Chalmers]{morrison2018large}
A.~Morrison, X.~Xiong, M.~Higgs, M.~Bell, and M.~Chalmers.
\newblock A large-scale study of iphone app launch behaviour.
\newblock In \emph{Proceedings of the 2018 CHI Conference on Human Factors in
  Computing Systems}, page 344. ACM, 2018.

\bibitem[Murnane et~al.(2016)Murnane, Abdullah, Matthews, Kay, Kientz,
  Choudhury, Gay, and Cosley]{murnane2016mobile}
E.~L. Murnane, S.~Abdullah, M.~Matthews, M.~Kay, J.~A. Kientz, T.~Choudhury,
  G.~Gay, and D.~Cosley.
\newblock Mobile manifestations of alertness: Connecting biological rhythms
  with patterns of smartphone app use.
\newblock In \emph{Proceedings of the 18th International Conference on
  Human-Computer Interaction with Mobile Devices and Services}, pages 465--477.
  ACM, 2016.

\bibitem[Singh and Ghosh(2017)]{Singh:2017:IIS:3171581.3134730}
V.~K. Singh and I.~Ghosh.
\newblock Inferring individual social capital automatically via phone logs.
\newblock \emph{Proc. ACM Hum.-Comput. Interact.}, 1\penalty0 (CSCW):\penalty0
  95:1--95:12, Dec. 2017.
\newblock ISSN 2573-0142.
\newblock \doi{10.1145/3134730}.
\newblock URL \url{http://doi.acm.org/10.1145/3134730}.

\bibitem[Wang et~al.(2015)Wang, Harari, Hao, Zhou, and
  Campbell]{wang2015smartgpa}
R.~Wang, G.~Harari, P.~Hao, X.~Zhou, and A.~T. Campbell.
\newblock Smartgpa: how smartphones can assess and predict academic performance
  of college students.
\newblock In \emph{Proceedings of the 2015 ACM international joint conference
  on pervasive and ubiquitous computing}, pages 295--306. ACM, 2015.

\bibitem[Wang et~al.(2018)Wang, Wang, daSilva, Huckins, Kelley, Heatherton, and
  Campbell]{wang2018tracking}
R.~Wang, W.~Wang, A.~daSilva, J.~F. Huckins, W.~M. Kelley, T.~F. Heatherton,
  and A.~T. Campbell.
\newblock Tracking depression dynamics in college students using mobile phone
  and wearable sensing.
\newblock \emph{Proceedings of the ACM on Interactive, Mobile, Wearable and
  Ubiquitous Technologies}, 2\penalty0 (1):\penalty0 43, 2018.

\bibitem[Zhao et~al.(2016)Zhao, Ramos, Tao, Jiang, Li, Wu, Pan, and
  Dey]{zhao2016discovering}
S.~Zhao, J.~Ramos, J.~Tao, Z.~Jiang, S.~Li, Z.~Wu, G.~Pan, and A.~K. Dey.
\newblock Discovering different kinds of smartphone users through their
  application usage behaviors.
\newblock In \emph{Proceedings of the 2016 ACM International Joint Conference
  on Pervasive and Ubiquitous Computing}, pages 498--509. ACM, 2016.

\end{thebibliography}

\end{document}